\documentclass[letterpaper, 10 pt, conference]{ieeeconf}  % Comment this line out if you need a4paper

\IEEEoverridecommandlockouts                              % This command is only needed if 
\usepackage{graphics} % for pdf, bitmapped graphics files
\usepackage{epsfig} % for postscript graphics files
\usepackage{times} % assumes new font selection scheme installed
\usepackage{amsmath} % assumes amsmath package installed
\usepackage{amssymb}  % assumes amsmath package installed
\usepackage{xcolor}
\usepackage{setspace}
\usepackage{hyperref}
\usepackage{subcaption}
\usepackage{algorithm}
\usepackage[noend]{algpseudocode}
\usepackage{xpatch}

\usepackage{todonotes}

\newcommand{\Ep}[2]{\mathop{\mathbb E_{#1}\/}\left\{{#2}\right\}}

\newcommand{\policy}{\pi_{\mathbf{\theta}}}

\newcommand{\policyp}[1]{\pi_{\mathbf{\theta}}\left(#1\right)}

\newcommand{\statet}[1]{\mathbf{x}_{#1}}

\newcommand{\actiont}[1]{\mathbf{u}_{#1}}

\newcommand{\argmin}[1]{\underset{#1}{\mathrm{arg\,min}\,}}

\newlength{\defaulttextfloatsep}
\newlength{\defaultintextsep}
\makeatletter
\algrenewcommand\ALG@beginalgorithmic{\small}
\makeatother
\makeatletter
\xpatchcmd{\algorithmic}{\itemsep\z@}{\itemsep=0.1em}{}{}
\makeatother

\title{\LARGE \bf
Synthesizing Neural Network Controllers with Probabilistic Model-Based Reinforcement Learning
}

\author{Juan Camilo Gamboa Higuera, David Meger, and Gregory Dudek
\thanks{The authors are part of the Center for Intelligent Machines and the
        School of Computer Science, McGill University, Montreal, Canada.
        This work was supported by NSERC through funding for the NSERC Canadian
        Field Robotics Network
        {\tt\small \{gamboa,dmeger,dudek\}@cim.mcgill.ca}}
}

\begin{document}

\maketitle
%%%%%%%%%%%%%%%%%%%%%%%%%%%%%%%%%%%%%%%%%%%%%%%%%%%%%%%%%%%%%%%%%%%%%%%%%%%%%%%%
\begin{abstract}
We present an algorithm for rapidly learning neural network policies for robotics systems. The algorithm follows the model-based reinforcement learning paradigm and improves upon existing algorithms: PILCO and a sample-based version of PILCO with neural network dynamics (Deep-PILCO). To improve convergence, we propose a model-based algorithm that uses fixed random numbers and clips gradients during optimization. We propose training a neural network dynamics model using variational dropout with truncated Log-Normal noise. These improvements enable data-efficient synthesis of complex neural network policies. We test our approach on a variety of benchmark tasks, demonstrating data-efficiency that is competitive with that of PILCO, while being able to optimize complex neural network controllers. Finally, we assess the performance of the algorithm for learning motor controllers for a six legged autonomous  underwater vehicle. This demonstrates the potential of the algorithm for scaling up the dimensionality and dataset sizes, in more complex tasks.
\end{abstract}
%%%%%%%%%%%%%%%%%%%%%%%%%%%%%%%%%%%%%%%%%%%%%%%%%%%%%%%%%%%%%%%%%%%%%%%%%%%%%%%%
\vspace{-0.25em}
\section{Introduction}
Model-based reinforcement learning (RL) is an attractive framework for addressing the synthesis of controllers for robots of all kinds due to its promise of data-efficiency. An RL agent can use learned dynamics models to search for good controllers in simulation. This has the potential of minimizing costly trials on real robots. Minimizing interactions, however, means that datasets will often not be large enough to obtain accurate models. Bayesian models are very helpful in this situation. Instead of requiring an accurate model, the robot agent may keep track of a distribution over hypotheses of models that are compatible with its experience. Evaluating a controller then involves quantifying its performance over the model distribution. To improve its chances of working in the real world an effective controller should perform well, on average, on models drawn from this distribution. PILCO (Probabilistic Inference and Learning for COntrol) and Deep-PILCO are successful applications of this idea. 

PILCO~\cite{deisenroth2015gaussian} uses Gaussian Process (GP) models to fit one-step dynamics and networks of radial basis functions (RBFs) as feedback policies. PILCO has been shown to perform very well with little data in simulated tasks and on real robots~\cite{deisenroth2015gaussian}. We have used PILCO successfully for synthesizing swimming controllers for an underwater swimming robot~\cite{meger2015learning}. However, PILCO is computationally expensive. Model fitting scales $O(Dn^3)$ and long-term predictions scale $O(D^3n^2)$, where $n$ is the dataset size and $D$ is the number of state dimensions, limiting its applicability only to scenarios with small datasets and low dimensionality.
\begin{figure}[t!]
    \centering
    \includegraphics[width=0.475\textwidth]{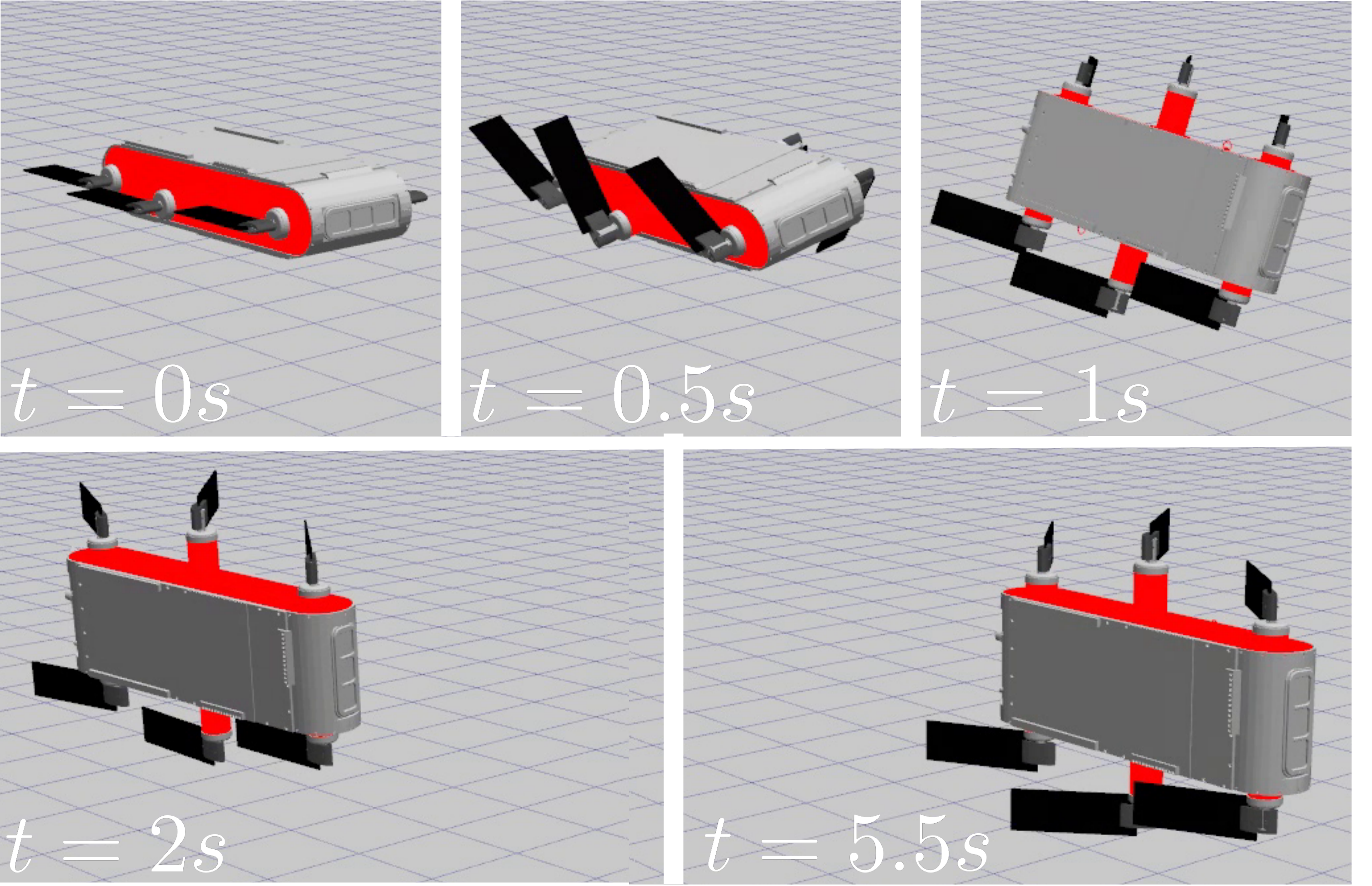}
    \caption{\small The AQUA robot executing a \emph{6-leg knife edge} maneuver. The robot starts in its resting position and must swim forward at a constant depth while stabilizing a roll angle of 90 degrees. The sequence of images illustrates a controller obtained with our proposed approach.}
    \label{fig:6_leg_kf}
\end{figure}
 
Deep-PILCO~\cite{gal2016improving} aims to address these limitations by employing Bayesian Neural Networks (BNNs), implemented via binary dropout~\cite{srivastava2014dropout,gal2016dropout}. Deep-PILCO performs a sampling-based procedure for simulating with BNN models of the dynamics. Policy search and model learning are done via stochastic gradient optimization, which scales more favorably to larger datasets and higher dimensionality. Deep-PILCO has been shown to result in better policies for a cart-pole swing-up benchmark task, but show reduced data efficiency when compared with PILCO. We extend on the results of~\cite{gal2016improving} by:
\begin{itemize}
    \item Modifying the simulation procedure to incorporate the use of fixed random numbers for policy optimization%~\cite{ng2000pegasus}
    \item Clipping gradients to stabilize optimization with back-propagation through time (BPTT)
    \item Using BNNs with multiplicative parameter noise where the noise distribution is adapted from data~\cite{lognormaldropout}% ( e.g. Truncated Log-Normal multiplicative noise~\cite{lognormaldropout})
\end{itemize}
We show how these improvements allow us to optimize neural network controllers with Deep-PILCO, while matching the data efficiency of PILCO on the cart-pole swing-up task; i.e. learning a successful controller with the same amount of experience. We also show how training stochastic policies (implemented as BNNs) can be beneficial for the convergence of robust policies. Finally, we demonstrate how these methods can be applied for learning swimming controllers for a 6 legged autonomous underwater vehicle.
 \vspace{-0.5em}
\section{Related Work}
Dynamics models have long been a core element in the modeling and control of robotic systems. Trajectory optimization approaches \cite{ddp,ilqr,tasaControlLimited} can produce highly effective controllers for complex robotic systems when precise analytical models are available. For complex and stochastic systems such as swimming robots, classical models are less reliable. In these cases, either performing online system identification \cite{softRobIJRR2015} or learning complete dynamics models from data has proven to be effective, and can be integrated tightly with model-based control schemes \cite{modelLearningSurvey,atkeson_lwr,heliRL,gps}. %TODO Juan, I think you have a more current set of references. Here would be a good place to talk about gp models building up to bayesian neural networks

Multiple works have applied Deep RL methods to learn various continuous control tasks~\cite{gu2016continuous,heess2015learning}, including full-body control of humanoid characters \cite{ddpg}. These methods do not assume a known reward function, estimating the value of each action from experience. Along with their model-free nature, this results in lower data efficiency compared with the methods we consider here, but there are ongoing efforts to connect model-based and model-free approaches \cite{MBMF}. 

The most similar works to our own are those which use probabilistic dynamics models for policy optimization. Locally linear controllers can be learned in this fashion, for example by extending the classical Differential Dynamic Programming (DDP) \cite{pddp} method or Iterative LQG \cite{GPILQG} to use GP models. For more complex robots, it is desirable to learn complex non-linear policies using the predictions of learned dynamics. Black-DROPS \cite{bbox_policy_search} has recently shown promising performance competitive with the gradient-based PILCO \cite{deisenroth2015gaussian} for training GP and NN policies using GP dynamics models. As yet, we are only aware of BNNs being used in the policy learning loop within Deep-PILCO \cite{gal2016improving}, which is the method we directly improve upon. Our approach is the first model-based RL approach to utilize BNNs for both the dynamics as well as the policy network. % TODO Juan please confirm my statements above
\vspace{-0.25em}
\section{Problem Statement}
We focus on model-based policy search methods for episodic tasks. We consider systems that can be modeled with discrete-time dynamics $\statet{t+1} = f(\statet{t}, \actiont{t})$, where $f$ is unknown, with states $\statet{t} \in \mathbb{R}^D$ and controls $\actiont{t} \in \mathbb{R}^U$, indexed by time-step $t$. The goal is to find the parameters $\theta$ of a policy $\policy$ that minimize a task-dependent cost function $c$ accumulated over a finite time horizon $H$,
\begin{align}
\begin{aligned}
    \argmin{\theta}\quad& J(\theta) = \Ep{\mathbf{\tau}}{\sum_{t=1}^H c(\statet{t}) \Bigm\vert \theta}.&\\
\end{aligned}
\label{eq:opt_prob}
\end{align}
The expectation in our case is due to not knowing the true dynamics $f$, which induces a distribution over trajectories $p(\tau) = p(\statet{1}, \actiont{1}, ... , \statet{H}, \actiont{H} \vert f)$. The objective could be minimized by black-box optimization or likelihood ratio methods, obtaining trajectory samples directly from the target system. However, such methods are known to require a large number of evaluations, which may be impractical for applications with real robot systems. An alternative is to  use experience to fit a model of the dynamics $f_\mathbf{w}$ and use it to estimate the objective in Eq.~(\ref{eq:opt_prob}). Alg.~(\ref{alg:mbrl}) describes a sketch for model-based optimization methods. A goal of these methods is \emph{data-efficiency}: to use as little real-world experience as possible. Since we consider fixed horizon tasks, data-efficiency can be measured in the number of episodes, or trials, until the task is successfully learned.
\begin{algorithm}
\algrenewcommand\algorithmicindent{1.5em}
\caption{Episodic Model-Based RL}
\label{alg:mbrl}
\begin{algorithmic}[1]
\State Initialize parameters $\theta$, $\mathbf{w}$ and dataset $\mathcal{D}$
\For {episode $e$ in $1 ... N_{\mathrm{trials}}$}
    \State Obtain $\tau_e$ by executing $\pi_{\theta}$ for $H$ steps on robot
    \State Append $\tau_e$ to $\mathcal{D}$
    \State Use $\mathcal{D}$ to update $\mathbf{w}$ \Comment{model learning}
    \State Use $f_\mathbf{w}$ to minimize $J(\theta)$, update $\theta$\Comment{policy optimization}
\EndFor
\State Return $\pi_\theta, f_\mathbf{w}$
\end{algorithmic}
\end{algorithm}
\vspace{-0.5em}
\section{Background}
\subsection{Learning a dynamics model with BNNs}
A key to data-efficiency is avoiding \emph{model bias}~\cite{atkeson1997,deisenroth2010efficient}, i.e. optimizing Eq.~(\ref{eq:opt_prob}) with a model that makes bad predictions with high confidence. BNNs address model bias by using the posterior distribution over their parameters. Given a model $f_\mathbf{w}$ with parameters $\mathbf{w}$ and a dataset $\mathcal{D} = \lbrace \mathbf{X}, \mathbf{Y} \rbrace$ we'd like to use the posterior $p(\mathbf{w} \lvert \mathcal{D})$ to make predictions at new test points. This distribution represents the uncertainty about the true value of $\mathbf{w}$, which induces uncertainty on the model predictions: $p(\mathbf{y}) = \int p(\mathbf{y} \vert f_\mathbf{w}, \mathbf{x}) p(\mathbf{w} \vert \mathcal{D}) d\mathbf{w} $, where $\mathbf{y}$ is the prediction at test point $\mathbf{x}$. Using the true posterior for predictions on a neural network is intractable. Fortunately, various methods based on variational inference exist, which use tractable approximate posteriors and Monte Carlo integration for predictions~\cite{gal2016dropout, blundell2015weight,kingma2015variational, molchanov2017variational, gal2017concrete}. Fitting is done by minimizing the Kullback-Leibler (KL) divergence between the true  and the approximate posterior, which can done by optimizing the objective
\begin{align}
\begin{aligned}
    \mathcal{L}(\mathbf{w}) = -\mathcal{L}_{\mathcal{D}}(\mathbf{w}) + D_{KL}\left(q(\mathbf{w})\vert p(\mathbf{w})\right) \\
\end{aligned}
\label{eq:model_loss}
\end{align}
where $\mathcal{L}_{\mathcal{D}}$ is the expected value of the likelihood $p\left(\mathcal{D}\vert \mathbf{w}\right)$, $q(\mathbf{w})$ is the approximate posterior and $p(\mathbf{w})$ is a user-defined prior on the parameters. These methods usually set $q(\mathbf{w})$ as a deterministic transformation of noise samples $\mathbf{w} = g(\mathbf{\psi}, \mathbf{z}_\mathbf{w}),\, \mathbf{z}_\mathbf{w} \sim p(\mathbf{z}_\mathbf{w})$, where $\mathbf{\psi}$ are the parameters of the posterior~\cite{kingma2015variational}. For example, in binary dropout $g$ multiplies the weights matrices for each layer of the network with the dropout masks $\mathbf{z}_\mathbf{w}$, consisting of diagonal noise matrices with entries drawn from a Bernoulli distribution with dropout probability $p$~\cite{srivastava2014dropout}. To fit the dynamics model, we build the dataset $\mathcal{D}$ of tuples $\langle (\statet{t}, \actiont{t}), \Delta_t  \rangle$; where $(\statet{t}, \actiont{t}) \in \mathbb{R}^{D+U}$ are the state-action pairs that we use as input to the dynamics model, and $\Delta_t = \statet{t} - \statet{t-1} \in \mathbb{R}^{X}$ are the changes in state after applying action $\actiont{t}$. We fit the model by minimizing the objective in Eq.(~\ref{eq:model_loss}) via stochastic gradient descent.
\vspace{-0.3em}
\subsection{Policy optimization with learned models}
%\begin{figure}
%    \centering
%    \includegraphics[width=0.495\textwidth]{figures/PILCO_module_as_RNN.pdf}
%    \caption{\small An overview diagram of the components used for sampling simulated trajectories. When both the policy and the model are implemented with neural networks, this model is effectively a simple recurrent neural network (RNN); with the same problems with long-term dependencies as traditional RNNs.}
%    \label{fig:pilco_module}
%    \vspace{-1em}
%\end{figure}
To estimate the objective function in Eq.~(\ref{eq:opt_prob}) we base our approach on Deep-PILCO~\cite{gal2016improving}, which we summarize in Alg.~(\ref{alg:deep_pilco}). For every optimization iteration, the algorithm draws particles consisting of an initial state and a set of weights sampled from $q(\mathbf{w})$, as shown in line 2. For the models used in \cite{gal2016improving} and this work, sampling weights is equivalent to sampling dropout masks $\mathbf{z}_\mathbf{w}$. The loop in lines 4 to 6 can be executed in parallel using batch processing. This algorithm requires the task cost function $c$ to be known and differentiable. Deep-PILCO uses back-propagation through time (BPTT) to estimate the policy gradients $\nabla_{\theta} J(\theta)$.
\begin{algorithm}[t!]
\algrenewcommand\algorithmicindent{1.5em}
\caption{Policy search with Deep-PILCO}
\label{alg:deep_pilco}
\begin{algorithmic}[1]
\For {$j$ in $1 ... N_{\mathrm{opt}}$}
    \State Sample $K$ particles $\lbrace(\statet{1}^{(k)}, f_\mathbf{w^{(k)}}) \vert 1\leq k \leq K\rbrace$
    \For {$t$ in $1 ... H$}
        \For {$k$ in $1 ... K$}
            \State Evaluate policy $\actiont{t}^{(k)} = \policyp{\statet{t}^{(k)}}$
            \State Propagate state $\statet{t+1}^{(k)} = f_\mathbf{w^{(k)}}\left(\statet{t}^{(k)}, \actiont{t}^{(k)}\right)$
        \EndFor
        \State Fit mean $\mu_{\statet{t+1}}$ and covariance $\Sigma_{\statet{t+1}}$
        \State Resample $\statet{t+1}^{(k)}$ from $\mathcal{N}(\mu_{\statet{t+1}}, \Sigma_{\statet{t+1}})$
    \EndFor
    \State Evaluate objective $J(\theta) = \frac{1}{K}\sum_{k=1}^K \sum_{t=1}^H c(\statet{t}^{(k)})$
    \State Compute gradient estimate $\nabla_{\theta} J(\theta)$
    \State Update $\theta$ by stochastic gradient descent step
\EndFor{}
\end{algorithmic}
\end{algorithm}
\section{Improvements to Deep-PILCO}
\label{sec:methods}
Here we describe the changes we have done to Deep-PILCO that were crucial for improving its data-efficiency and obtaining the results we describe in Sec.~\ref{sec:results}. Our changes are summarized in Alg.~(\ref{alg:ours}).
This algorithm can still be executed efficiently using batch processing with state-of-the art deep learning frameworks.
\subsection{Common random numbers for policy evaluation}
The convergence of Algorithms~(\ref{alg:deep_pilco})~and~(\ref{alg:ours}) is highly dependent on the variance of the estimated gradient $\nabla_{\theta} J(\theta))$. In this case, the variance of the gradients is dependent on the sources of randomness for simulating trajectories: the initial state samples  $\statet{1}$, the multiplicative noise masks $\mathbf{z}_{\mathbf{w}}^{(k)}, \mathbf{z}_{\mathbf{\theta}}^{(k)}$, and the random numbers used for re-sampling $\mathbf{z}_t^{(k)}$. A common variance reduction technique used in stochastic optimization is to fix random numbers during optimization~\cite{kleinman1999simulation}. Using common random numbers (CRNs) reduces variance in two ways: gradient evaluations become deterministic and evaluations over different values for the optimization variable become correlated. We introduce CRNs by drawing all the random numbers we need for simulating trajectories at the beginning of the policy optimization (lines 1 to 3 in Alg.~(\ref{alg:ours}))  and keeping them fixed as the policy parameters are updated. This is possible because we use BNNs that rely on the re-parametrization trick~\cite{kingma2015variational} for evaluation. This is effective in reducing variance and improving convergence, but it may introduce bias. A simple way to deal with bias is to increase the number of particles $K$ used for gradient evaluation. We increased $K$ from 10, the number used in~\cite{gal2016improving}, to 100 for our experiments, and found it to improve convergence with small penalty on running time.

Fixing random numbers in the context of policy search is known as the PEGASUS\footnote{Policy Evaluation-of-Goodness And Search Using Scenarios} algorithm~\cite{ng2000pegasus}. PEGASUS consists of transforming a given Markov Decision Process (MDP) into "an equivalent one where all transitions are deterministic" by assuming access to a \emph{deterministic simulative model} of the MDP. A deterministic simulative model is one that has no internal random number generator, so any random numbers that are needed must be given to it as input. This is the case when using BNNs models. PEGASUS provides theoretical justification to our approach, particularly in that to decrease the upper bound on the error of estimates of $J(\theta)$ using CRNs it suffices to increase $K$.
\subsection{Stabilization for back-propagation through time}
\label{sec:clip}
As noted in~\cite{gal2016improving}, the recurrent application of BNNs in Algortihm~\ref{alg:deep_pilco} can be interpreted as a Recurrent Neural Network (RNN) model. As such, Deep-PILCO is prone to suffer from vanishing and exploding gradients when computing them via BPTT~\cite{pascanu2013difficulty}, especially when dealing with tasks that require long time horizon or very deep models for the dynamics and policy. Although numerous techniques have been proposed in the RNN literature, we opted to deal with these problems by using ReLU activations for the policy and dynamics model, and clipping the gradients to have a norm of at most $\nu$. We show the effect of various settings of the clipping value $\nu$ on the convergence of policy search in Fig.~(\ref{fig:cp_clip_comparison}) .
\begin{algorithm}[t!]
\algrenewcommand\algorithmicindent{1.5em}
\caption{Our method: Deep-PILCO with PEGASUS evaluation and gradient clipping}
\label{alg:ours}
\begin{algorithmic}[1]
\State Sample noise for dynamics $\lbrace \mathbf{z}_{\mathbf{w}}^{(k)} \vert 1\leq k \leq K\rbrace$
\State Sample noise for policy $\lbrace \mathbf{z}_{\theta}^{(k)} \vert 1\leq k \leq K\rbrace$
\State Sample state noise $\lbrace \mathbf{z}_t^{(k)} \vert 1\leq k \leq K, 1 \leq t \leq H\rbrace$
\For {$j$ in $1 ... N_{\mathrm{opt}}$}
    \State Sample $K$ particles $\lbrace \statet{1}^{(k)} \vert 1\leq k \leq K\rbrace$
    \For {$t$ in $1 ... H$}
        \For {$k$ in $1 ... K$}
            \State $\theta^{(k)} = g_1(\theta,\mathbf{z}_{\theta}^{(k)})$
            \State $\mathbf{w}^{(k)} = g_2(\mathbf{w},\mathbf{z}_{\mathbf{w}}^{(k)})$
            \State Evaluate policy $\actiont{t}^{(k)} = \pi_{\theta^{(k)}}\left(\statet{t}^{(k)}\right)$
            \State Propagate state $\statet{t+1}^{(k)} = f_\mathbf{w^{(k)}}\left(\statet{t}^{(k)}, \actiont{t}^{(k)}\right)$
        \EndFor
        \State Fit mean $\mu_{\statet{t+1}}$ and covariance $\Sigma_{\statet{t+1}}$
        \For {$k$ in $1 ... K$}
            \State $\statet{t+1}^{(k)}$ = $\mu_{\statet{t+1}} + \Sigma_{\statet{t+1}}\mathbf{z}_t^{(k)}$
        \EndFor
    \EndFor
    \State Evaluate objective $J(\theta) = \frac{1}{K}\sum_{k=1}^K \sum_{t=1}^H c(\statet{t}^{(k)})$
    \State Compute gradient estimate $\nabla_{\theta} J(\theta)$
    \If{$\nabla_{\theta} J(\theta) > \nu$}
        \State $\nabla_{\theta} J(\theta) \leftarrow \nu \frac{\nabla_{\theta} J(\theta)}{||\nabla_{\theta} J(\theta)||}$
    \EndIf
    \State Update $\theta$ by stochastic gradient descent step
\EndFor{}
\end{algorithmic}
\end{algorithm}
\subsection{BNN models with Log-Normal multiplicative noise}
We focused on methods that use multiplicative noise on the activations (e.g. binary dropout) because of their simplicity and computational efficiency. Deep-PILCO with binary dropout requires tuning the dropout probability to a value appropriate for the model size. We experimented with various BNN models~\cite{lognormaldropout,kingma2015variational,molchanov2017variational,gal2017concrete} to enable learning the dropout probabilities from data. The best performing method in our experiments was using truncated Log-Normal dropout with a truncated log-uniform prior $\mathrm{Log}\mathrm{U}_{[-10,0]}$. This choice prior causes the multiplicative noise $\mathbf{z}_{\mathbf{w}}$ to be constrained to values between 0 and 1~\cite{lognormaldropout}.
\subsection{Training neural network controllers}
While Deep-PILCO had been limited to training single-layer Radial Basis Function policies, the application of gradient clipping and CRNs allows stable training of deep neural network policies, opening the door for richer behaviors. We found that adding dropout to the policy networks improves performance. During policy evaluation, we sample policies the same way as we do for dynamics models: a policy sample corresponds to a set of dropout masks $\mathbf{z}_{\theta}^{(k)}$. Thus each simulated state particle has a corresponding dynamics model and policy, which remain fixed during policy optimization. This can be interpreted as attempting to learn a distribution of controllers that are likely to perform well over plausible dynamics models. We make the policy stochastic during execution on the target system by re-sampling the policy dropout mask $\mathbf{z}_{\theta}$ at every time step. This provides some amount of exploration that we found beneficial.%, in particular, for the double cart-pole swing-up task.
\section{Results}
\label{sec:results}
\begin{figure}[t!]
 \centering
    \begin{subfigure}[b]{0.18\textwidth}
        \includegraphics[width=\textwidth]{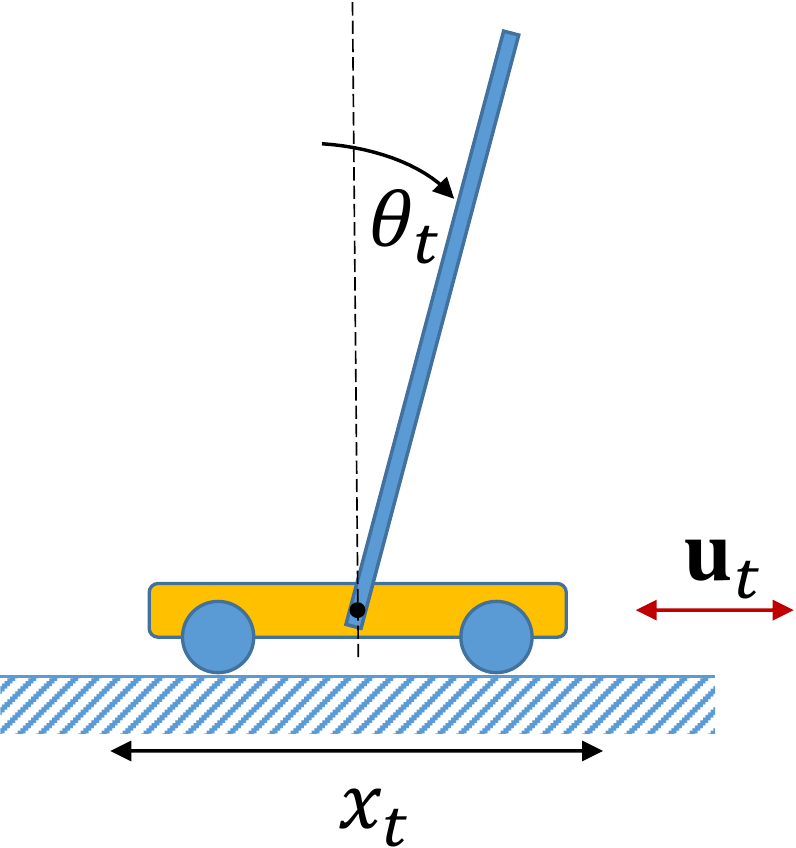}
        \caption{Cart-pole task}
        \label{fig:cpa}
    \end{subfigure}
    \begin{subfigure}[b]{0.18\textwidth}
        \includegraphics[width=\textwidth]{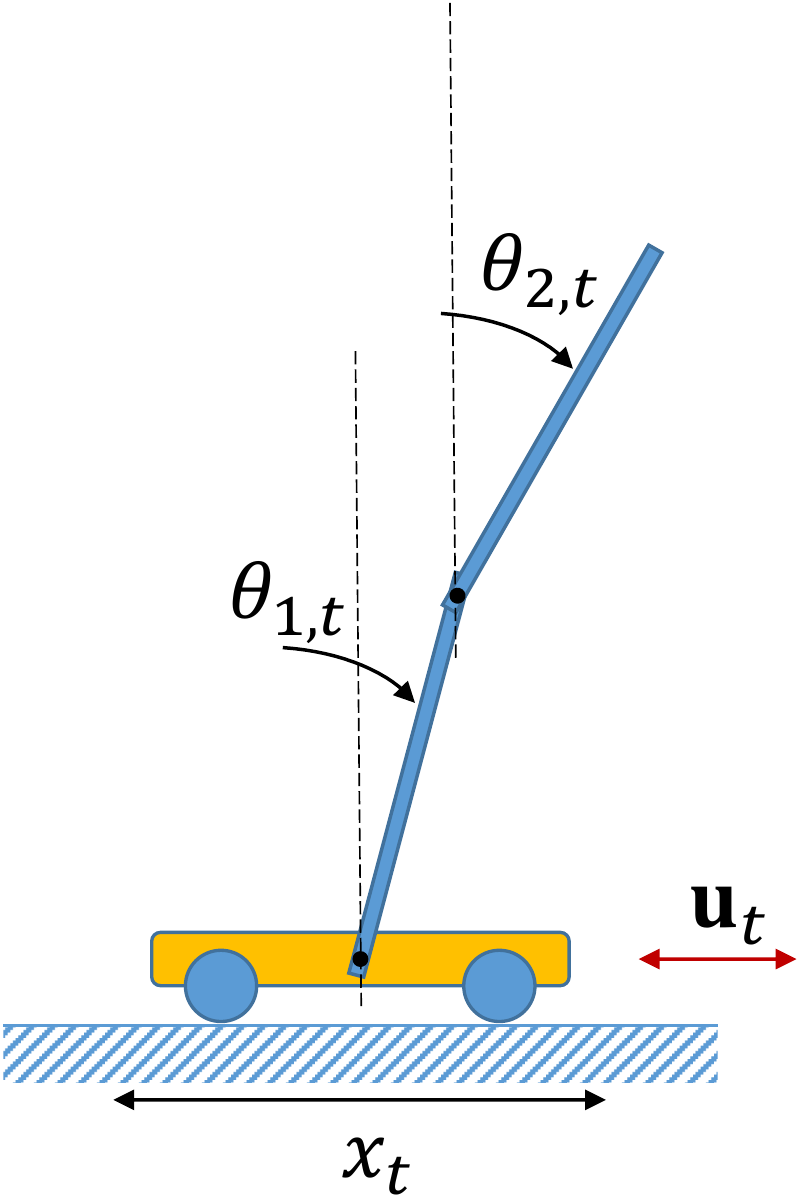}
        \caption{Double cart-pole task}
        \label{fig:dcpb}
    \end{subfigure}
    \caption{\small Benchmark tasks used in Sec.~\ref{sec:results}. In both tasks, the tip of the pendulum starts downright, with the cart centered at $x=0$. The goal is to balance the tip of the pole at its highest possible location, while keeping the cart at $x=0$. This occurs when $\theta=0$ for the cart-pole, and when $\theta_1=0, \theta_2=0$ in the double cart-pole.}
    \label{fig:cp_tasks}
\end{figure}
 We tested the improvements, described in Section~\ref{sec:methods}, on two benchmark scenarios: swinging up and stabilizing an inverted pendulum on a cart, and swinging up and stabilizing a double pendulum on a cart (see Fig.~(\ref{fig:cp_tasks})). The first task was meant to compare performance on the same experiment as~\cite{gal2016improving}. We chose the second scenario to compare the methods with a harder long-term prediction task; due to the chaotic dynamics of the double-pendulum. In both cases, the system is controlled by applying a horizontal force $\mathbf{u}$ to the cart. We also evaluate our approach on the gait learning tasks for an underwater hexapod robot~\cite{meger2015learning} to demonstrate the applicability of our approach for locomotion tasks on complex robot systems. We use the ADAM optimizer~\cite{DBLP:journals/corr/KingmaB14} for model fitting and policy optimization, with the default parameters suggested by the authors, and report the best results obtained after manual hyper-parameter tuning.
  \subsection{Cart-pole swing-up task}
\label{sec:cp_exps}
\begin{figure}[t!]
\vspace{0.1em}
    \centering
    \begin{subfigure}[b]{0.495\textwidth}
        \includegraphics[width=\textwidth]{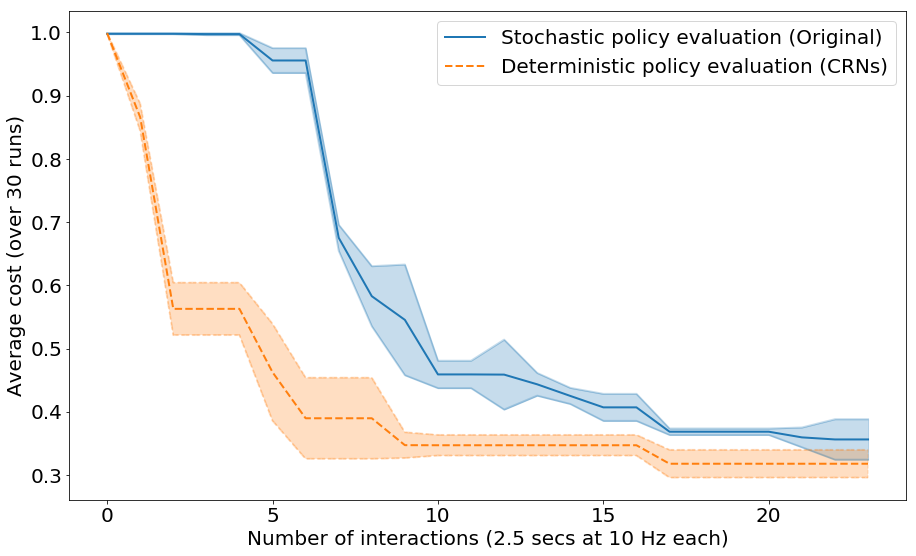}
        \caption{Comparison of the effect of introducing CRNs}
        \label{fig:crn_comparison}
    \end{subfigure}
    \begin{subfigure}[b]{0.35\textwidth}
        \includegraphics[width=\textwidth]{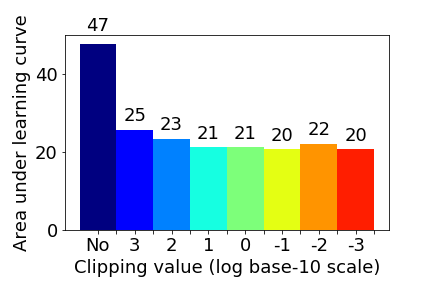}
        \vspace{-1.75em}
        \caption{Effect of clipping gradients}
        \label{fig:cp_clip_comparison}
    \end{subfigure}
    \caption{\small (a) Illustrates the benefit of fixing random numbers for policy evaluation (Alg.~(\ref{alg:ours})) vs. stochastic policy evaluations (Alg.~(\ref{alg:deep_pilco})). In (b) we show the area under the learning curve for the cart-pole task for various gradient clipping values (lower is better).}
    \label{fig:our_comparison}
\end{figure}
While previous experiments combining PILCO with PEGASUS were unsuccessful~\cite{deisenroth2010efficient}, we found its application to Deep-PILCO to result in a significant improvement on convergence when training neural network policies. Fig.~(\ref{fig:crn_comparison}) shows how the use of CRNs (Deterministic policy evaluation) results in faster convergence than the original Deep-PILCO formulation (Stochastic policy evaluation), which only matches the cost of our approach after around the 20th trial. These experiments were done with a learning rate of $10^{-4}$ and clipping value $\nu=1.0$.
Fig.~(\ref{fig:cp_clip_comparison}) illustrates the effect of gradient clipping for different values of $\nu$ for Alg.~(\ref{alg:ours}). The area under the learning curve gives us an idea of the speed of convergence as the clipping value changes. The trend is that any value of gradient clipping made a large improvement over not clipping at all and that the specific choice of clipping values was highly stable.
\begin{figure}[t!]
\vspace{0.1em}
 \centering
    \begin{subfigure}[b]{0.495\textwidth}
        \includegraphics[width=\textwidth]{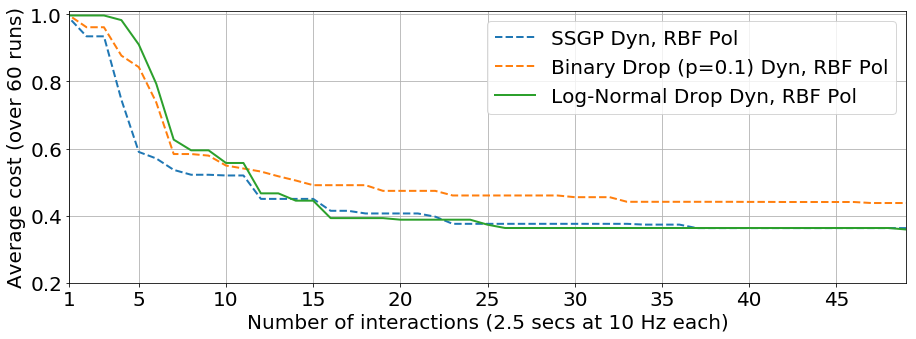}
        \caption{Cart-pole RBF Policies}
        \label{fig:cp_rbf_comparison}
    \end{subfigure}
    \begin{subfigure}[b]{0.495\textwidth}
        \includegraphics[width=\textwidth]{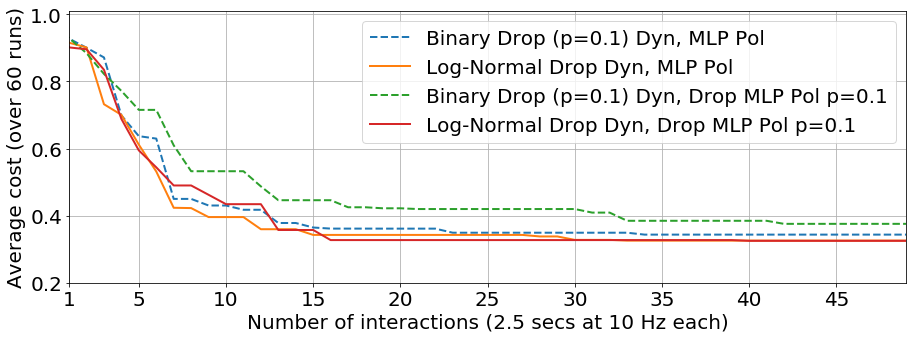}
        \caption{Cart-pole Deep Policies}
        \label{fig:cp_mlp_comparison}
    \end{subfigure}
    \caption{\small Cost per trial on the cart-pole swing-up task. In (a), we compare different dynamics models for learning RBF policies. (b) compares BNN models for learning NN policies, showing how our approach matches the data-efficiency of PILCO with better final performance.}
    \label{fig:cp_comparison}
\end{figure}

Fig.~(\ref{fig:cp_comparison}) summarizes our results for the cart-pole domain\footnote{The code used in these experiments is available at \url{https://github.com/juancamilog/kusanagi}.}. Fig.~(\ref{fig:cp_rbf_comparison}) illustrates the difference in performace between PILCO using sparse spectrum GP (SSGP) regression~\cite{quia2010sparse} for the dynamics and two versions of Deep-PILCO using BNN dynamics: one using binary dropout with dropout probability $p=0.1$, and the other using Log-Normal dropout with a truncated log-uniform  $\mathrm{Log}\mathrm{U}_{[-10,0]}$. 
The BNN models are ReLU networks with two hidden layers of 200 units and a linear output layer. The models predict heteroscedastic noise, which is used to corrupt the input to the policy during simulation. We used data from all previous episodes for model learning after each trial. The initial experience was gathered with a single execution of a policy that selects actions uniformly-at-random. The learning rate was set to $10^{-4}$ for model learning and $10^{-3}$ for policy optimization. The policies were RBF networks with 30 units. Fig.~(\ref{fig:cp_mlp_comparison}) provides a comparison of different  BNN dynamics models when training neural network policies. The policy networks are ReLU networks with two hidden layers of 200 units. For BNN policies (Drop MLP) we set a constant dropout probability $p=0.1$. Note that our method is able to train neural network controllers with better performance (lower cost) than either PILCO or Deep-PILCO with RBF controllers, within a similar number of trials. Using truncated Log-Normal dropout (Log-Normal Drop Dyn) for learning a stochastic policy (Drop MLP) results in the best performance for the cart-pole task.
 % TODO this would be a lot cleaner if we separated the experimental details like shared code and method for collecting random experience from the analysis 
\subsection{Double pendulum on cart swing-up task}
\begin{figure}[t!]
\vspace{0.1em}
    \centering
    \includegraphics[width=0.495\textwidth]{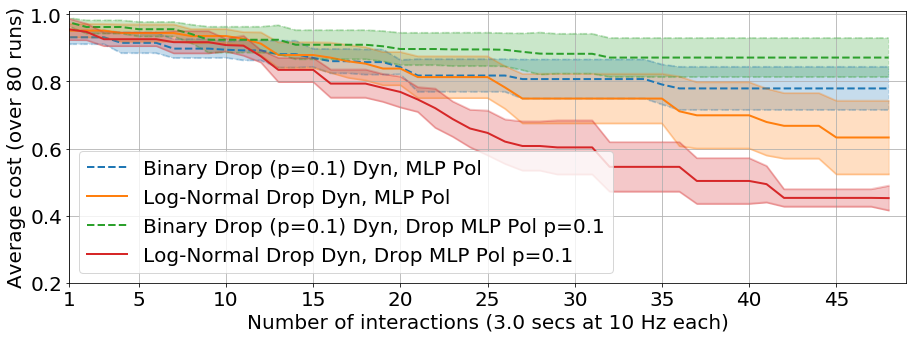}
    \caption{\small Cost per trial on the double cart-pole swing-up task. The shaded regions correspond to half a standard deviation. This demonstrates the benefit of using Log-Normal multiplicative noise for the dynamics with dropout regularization for the policies}
    \label{fig:dcp_comparison}
\end{figure}

Fig.~(\ref{fig:dcp_comparison}) illustrates the effect of learning neural network controllers on the more complicated double cart-pole swing-up task.  We were unable to get Alg.~(\ref{alg:deep_pilco}) with RBF policies to converge in this task. The setup is similar to the cart-pole task, but we change the network architectures as the dynamics are more complex. The dynamics models are ReLU networks with 4 hidden layers of 200 units and a linear output layer. The policies are ReLU networks with four hidden layers of 50 units. The learning rate for policy learning was set to $10^{-4}$.  The initial experience was comes from 2 runs with random actions. Here the differences in performance are more pronounced: our method converges after 42 trials, corresponding to 126 s of experience at 10 Hz. This is close to the 84 s at 13.3 Hz reported in~\cite{deisenroth2010efficient}. We see that the combination of BNN dynamics (Log-Normal Drop Dyn) and a BNN policy (Drop MLP Pol) results in the least number of trials for achieving the lowest cost.
\subsection{Learning swimming gaits on an underwater robot.}
\begin{figure*}[t!]
\vspace{0.4em}
    \centering
    \begin{subfigure}[c]{\textwidth}
        \hfill
        \begin{subfigure}[c]{0.495\textwidth}
            \includegraphics[width=\textwidth]{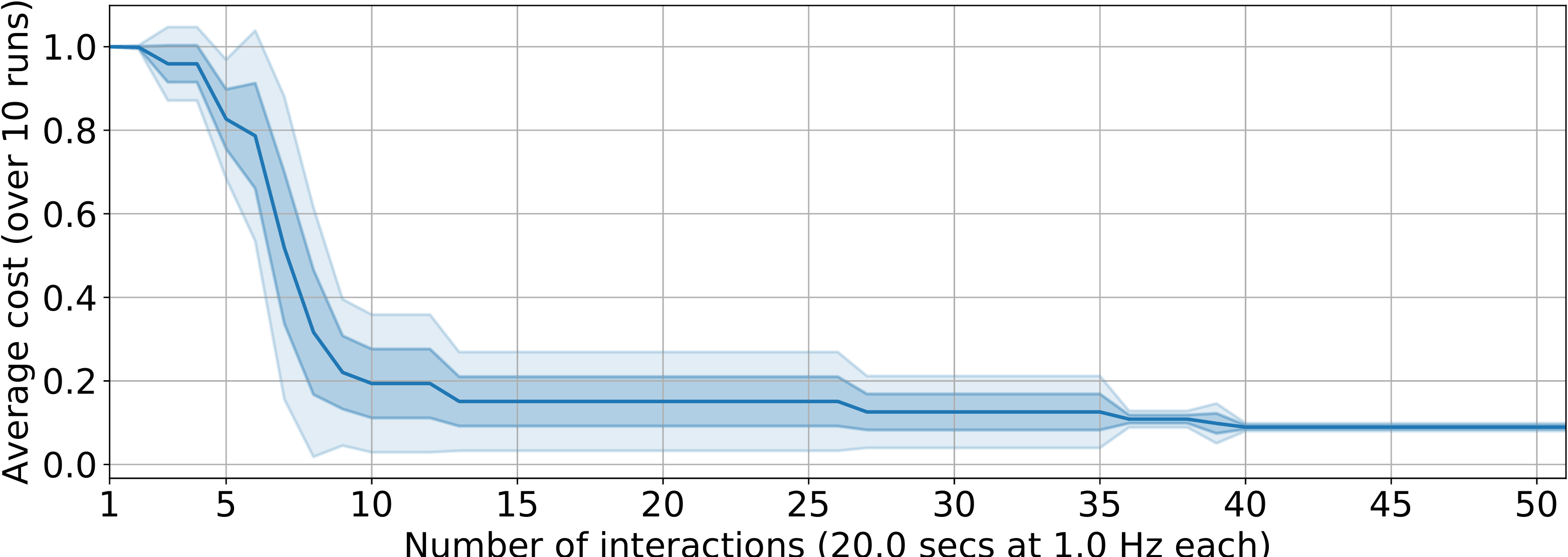}
        \end{subfigure}
        \hfill
        \begin{subfigure}[c]{0.495\textwidth}
            \includegraphics[width=\textwidth]{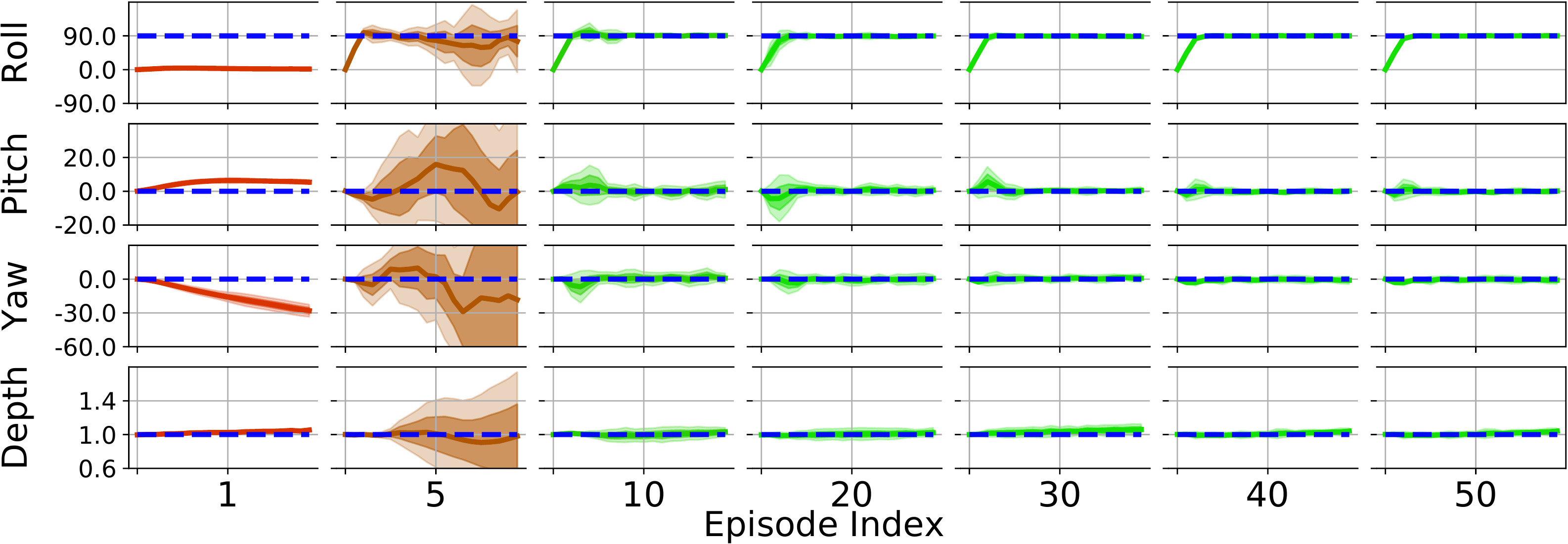}
        \end{subfigure}
        \hfill
        \vspace{-0.25em}
        \caption{\small 2-Leg Knife edge}
        \vspace{1em}
    \end{subfigure}
    \begin{subfigure}[c]{\textwidth}
        \hfill
        \begin{subfigure}[c]{0.495\textwidth}
            \includegraphics[width=\textwidth]{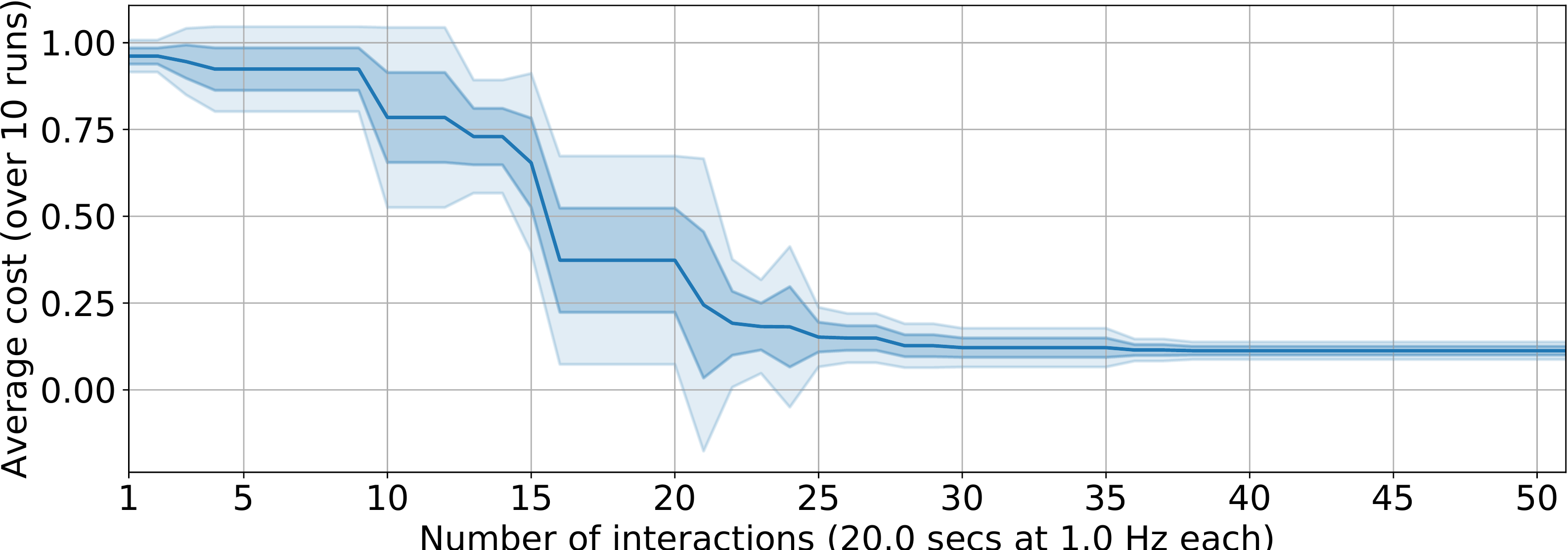}
        \end{subfigure}
        \hfill
        \begin{subfigure}[c]{0.495\textwidth}
            \includegraphics[width=\textwidth]{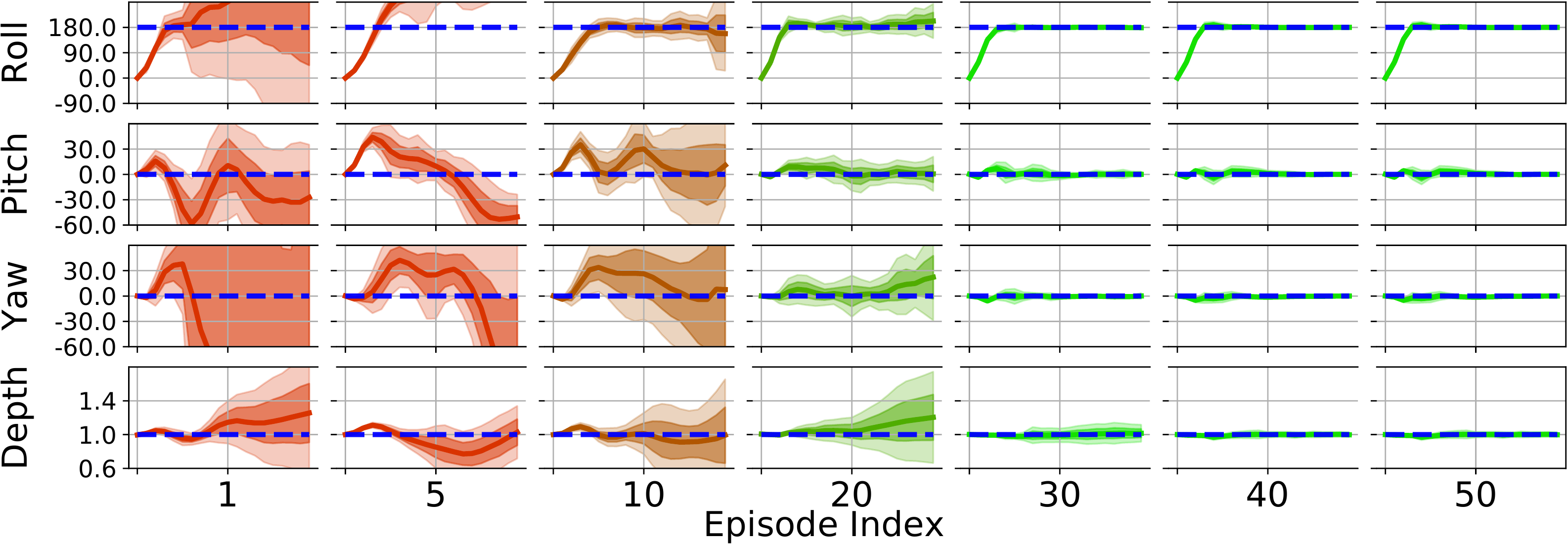}
        \end{subfigure}
        \hfill
        \vspace{-0.25em}
        \caption{\small 2-Leg Belly up}
        \vspace{1em}
    \end{subfigure}
    \begin{subfigure}[c]{\textwidth}
        \hfill
        \begin{subfigure}[c]{0.495\textwidth}
            \includegraphics[width=\textwidth]{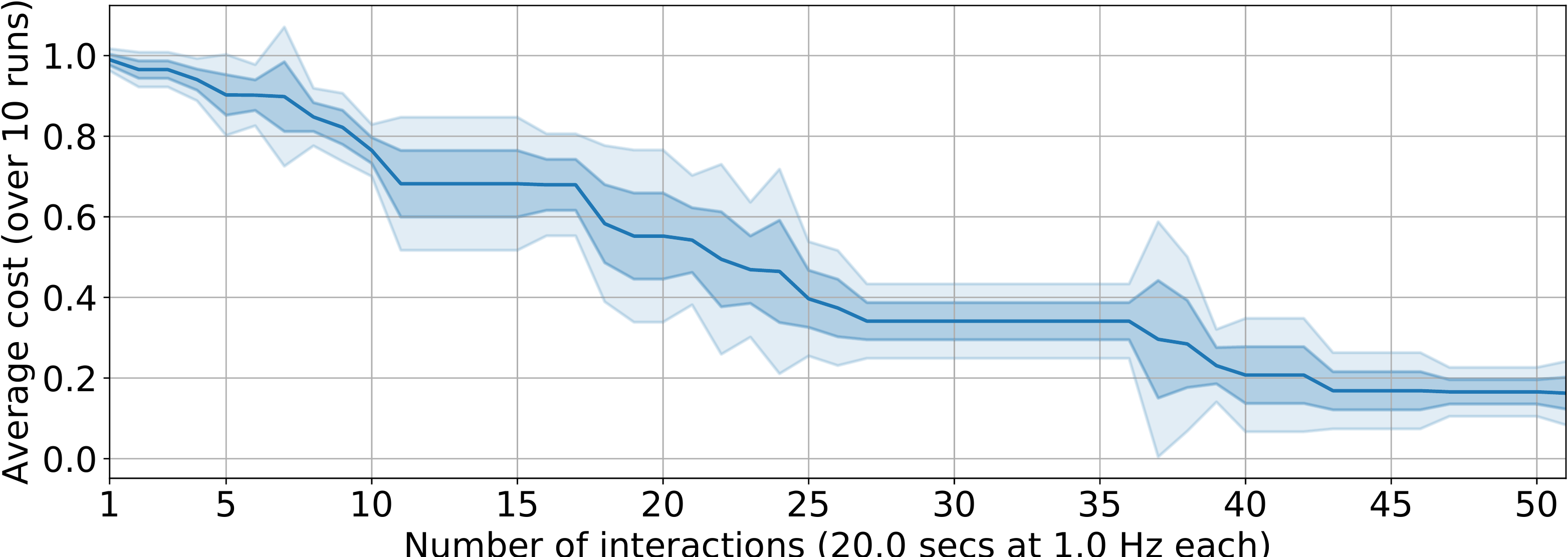}
        \end{subfigure}
        \hfill
        \begin{subfigure}[c]{0.495\textwidth}
            \includegraphics[width=\textwidth]{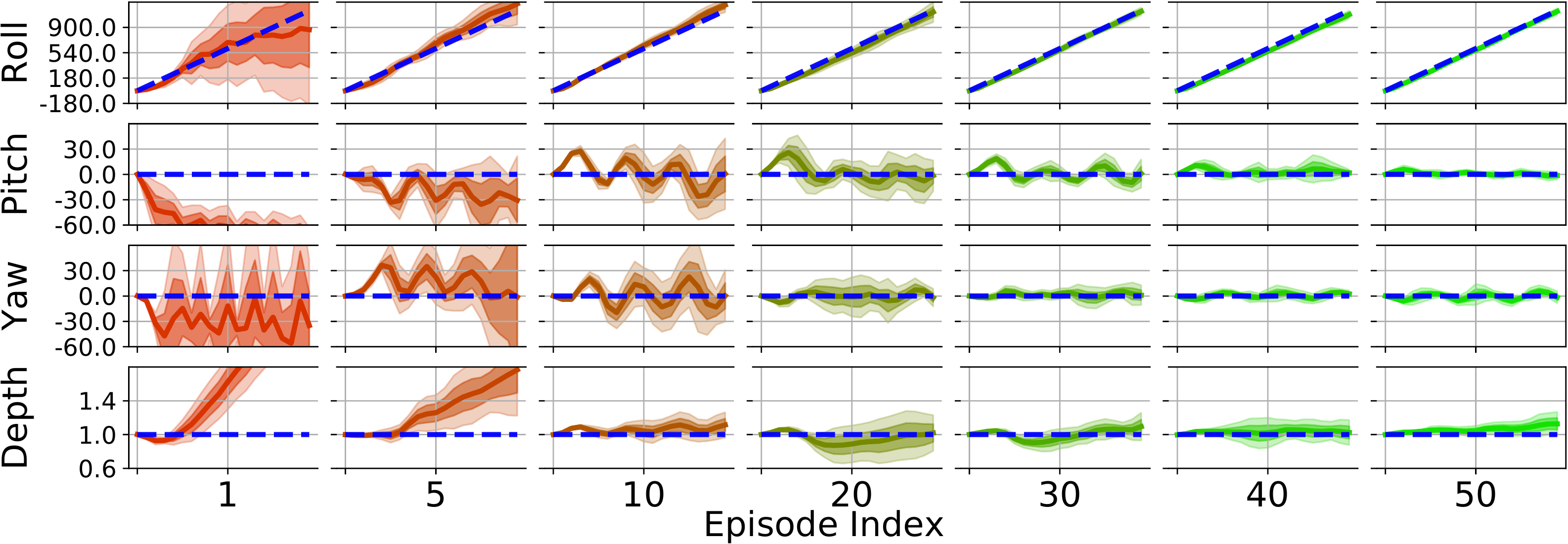}
        \end{subfigure}
        \hfill
        \vspace{-0.25em}
        \caption{\small 2-Leg Corkscrew}
        \vspace{1em}
    \end{subfigure}
    \begin{subfigure}[c]{\textwidth}
        \hfill
        \begin{subfigure}[c]{0.495\textwidth}
            \includegraphics[width=\textwidth]{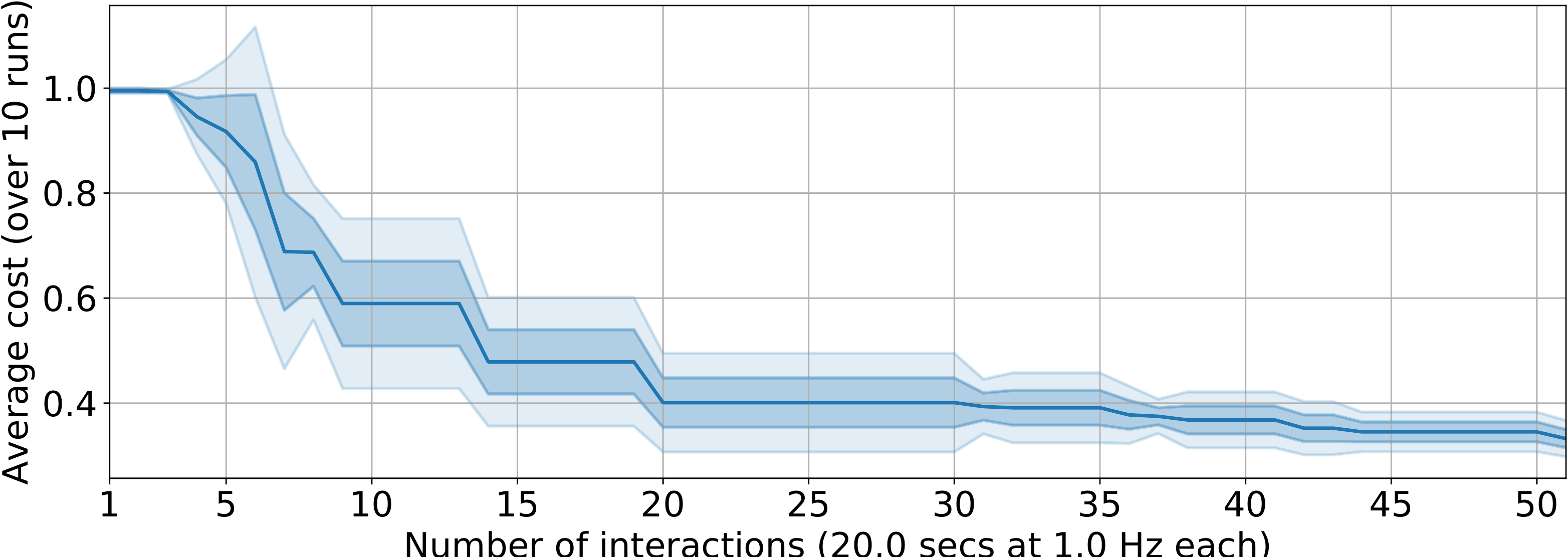}
        \end{subfigure}
        \hfill
        \begin{subfigure}[c]{0.495\textwidth}
            \includegraphics[width=\textwidth]{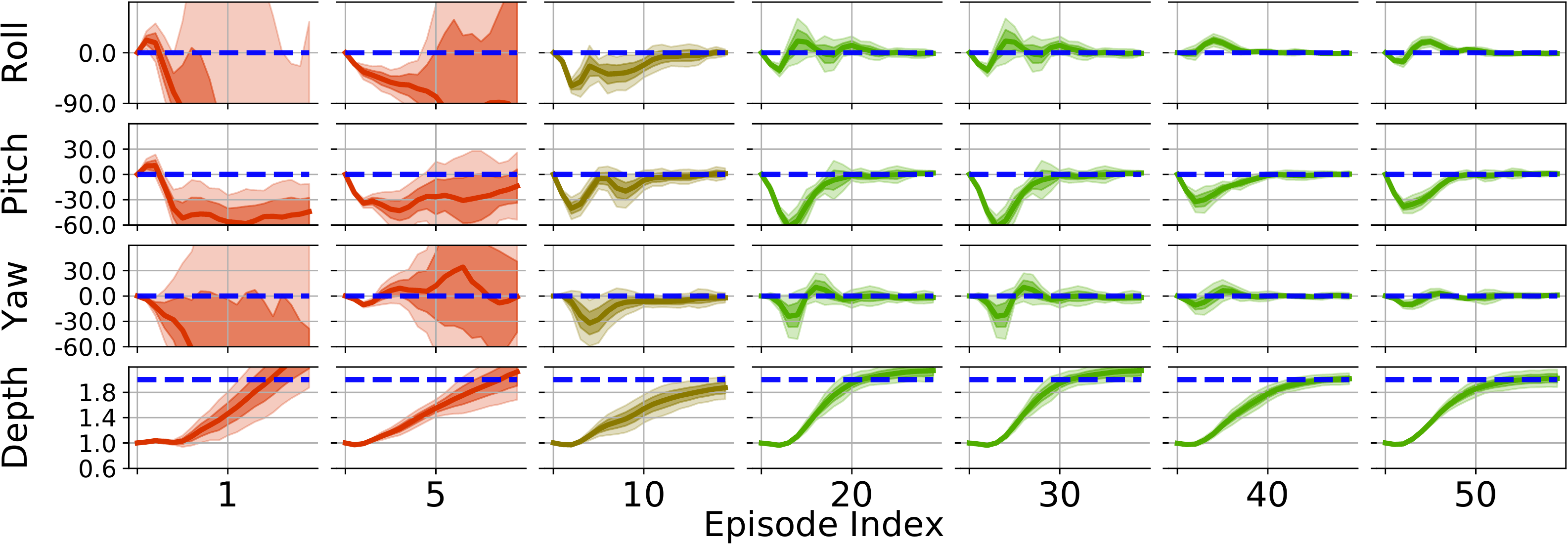}
        \end{subfigure}
        \hfill
        \vspace{-0.25em}
        \caption{\small 2-Leg Depth change}
        \vspace{1em}
    \end{subfigure}
    \vspace{-1em}
    \caption{\small Learning curve and the evolution of the trajectory distribution as learning progresses for 2-leg tasks. The robot learns to control its pose by setting the appropriate amplitudes and leg offset angles for its back 2 legs. The dashed lines represent the desired target states. Additional results and videos of these behaviours available at \url{https://github.com/mcgillmrl/robot_learning}}
    \label{fig:aqua2leg}
    \vspace{-1.5em}
\end{figure*}
 These tasks consist of finding feedback controllers for controlling the robot's 3D pose via periodic motion of its legs. Fig.~(\ref{fig:6_leg_kf}) illustrates the execution of a gait learned using our methods. The robot's state space consists of readings from its inertial measurement unit (IMU), its depth sensor and motor encoders. To compare with previously published results, the action space is defined as the parameters of the periodic leg command (PLC) pattern generator~\cite{meger2015learning}, with the same constraints as prior work. We conducted experiments on the following control tasks\footnote{The code used for these experiments and video examples of other learned gaits are available at \url{https://github.com/mcgillmrl/robot_learning}.}:
\begin{enumerate}
    \item \emph{knife edge}: Swimming straight-ahead with $90\deg$ roll
    \item \emph{belly up}: Swimming straight-ahead with $180\deg$ roll
    \item \emph{corkscrew}: Swimming straight-ahead with $120\deg$ rolling velocity (anti-clockwise)
    \item \emph{1 m depth change}: Diving and stabilizing 1 meter below current depth.
\end{enumerate}
There were two versions of these experiments. In the first one, which we call \emph{2-leg tasks}, the robot controls only the amplitudes and offsets of the two back legs (4 control dimensions). Its state corresponds to the angles and angular velocities, as measured by the IMU, and the depth sensor measurement (7 state dimensions). In the second version, the robot controls amplitudes and offsets and phases for all 6 legs (18 control dimensions). In this case, the state consists of the IMU and depth sensor readings plus the leg angles as measured form the motor encoders (13 state dimensions). We transform angle dimensions into their complex representation before passing the state as input to the dynamics model and policy, as described in~\cite{deisenroth2010efficient}. 
\begin{figure*}[t!]
\vspace{0.4em}
    \centering
    \begin{subfigure}[c]{\textwidth}
        \hfill
        \begin{subfigure}[c]{0.495\textwidth}
            \includegraphics[width=\textwidth]{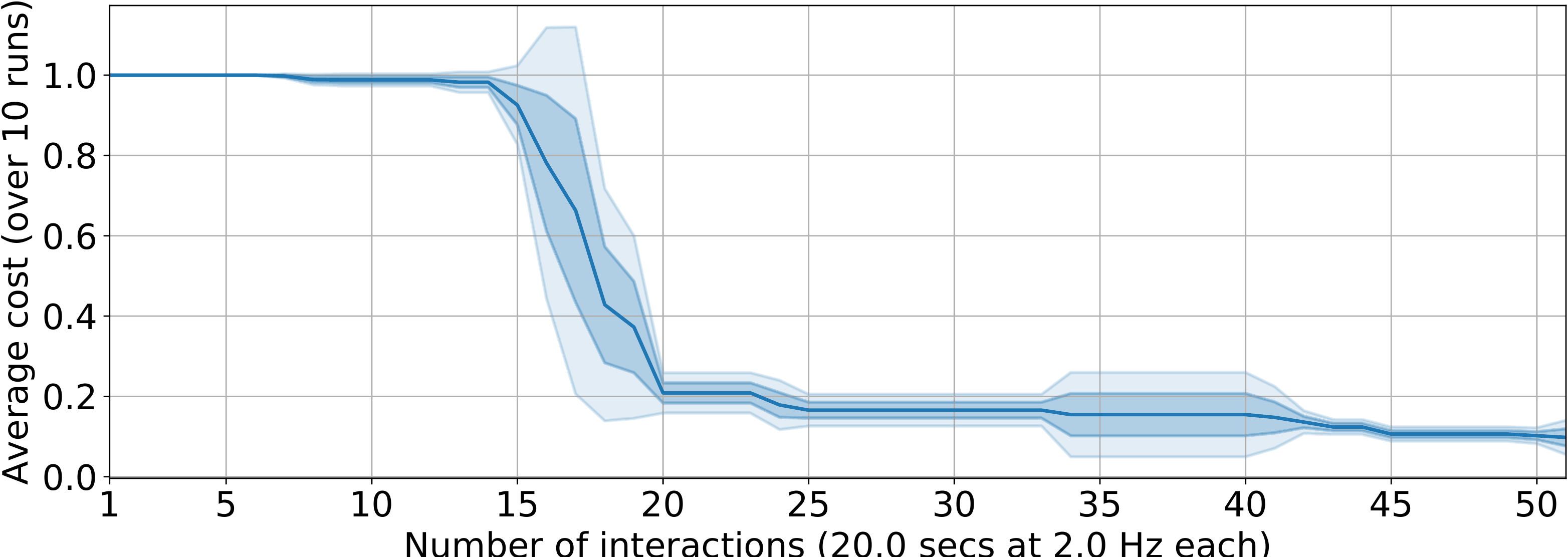}
        \end{subfigure}
        \hfill
        \begin{subfigure}[c]{0.495\textwidth}
            \includegraphics[width=\textwidth]{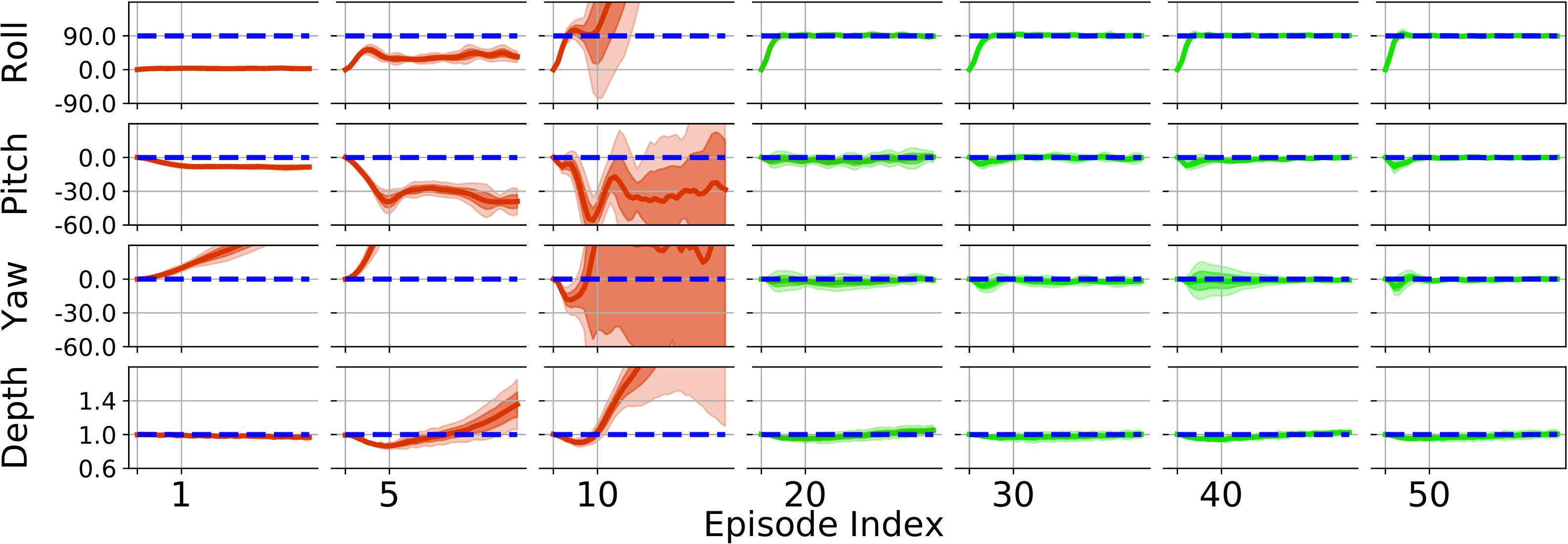}
        \end{subfigure}
        \hfill
        \vspace{-0.25em}
        \caption{\small 6-Leg Knife edge}
        \vspace{1em}
    \end{subfigure}
    \begin{subfigure}[c]{\textwidth}
        \hfill
        \begin{subfigure}[c]{0.495\textwidth}
            \includegraphics[width=\textwidth]{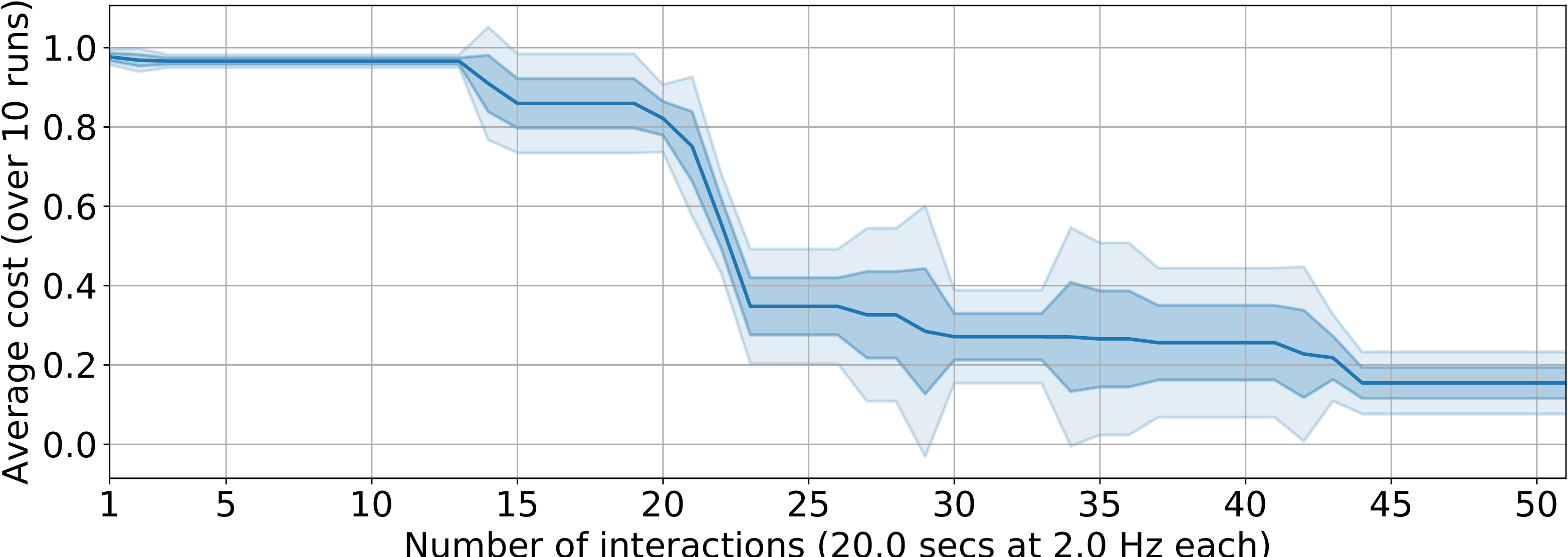}
        \end{subfigure}
        \hfill
        \begin{subfigure}[c]{0.495\textwidth}
            \includegraphics[width=\textwidth]{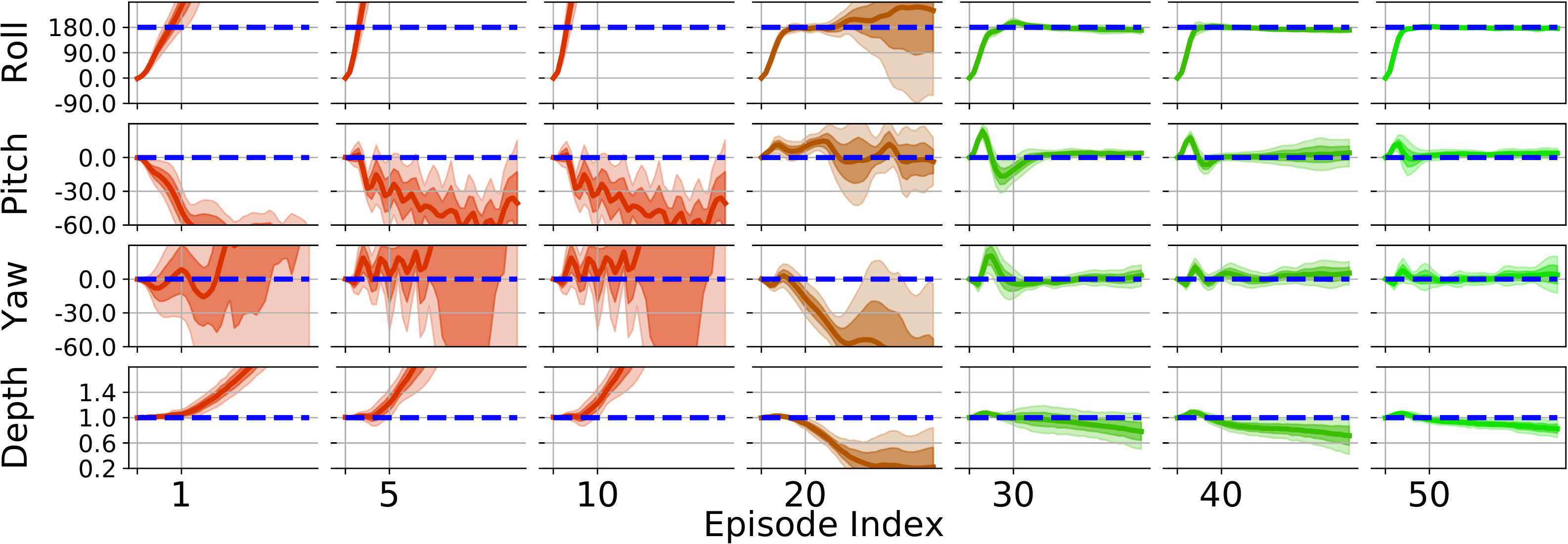}
        \end{subfigure}
        \hfill
        \vspace{-0.25em}
        \caption{\small 6-Leg Belly up}
        \vspace{1em}
    \end{subfigure}
    \begin{subfigure}[c]{\textwidth}
        \hfill
        \begin{subfigure}[c]{0.495\textwidth}
            \includegraphics[width=\textwidth]{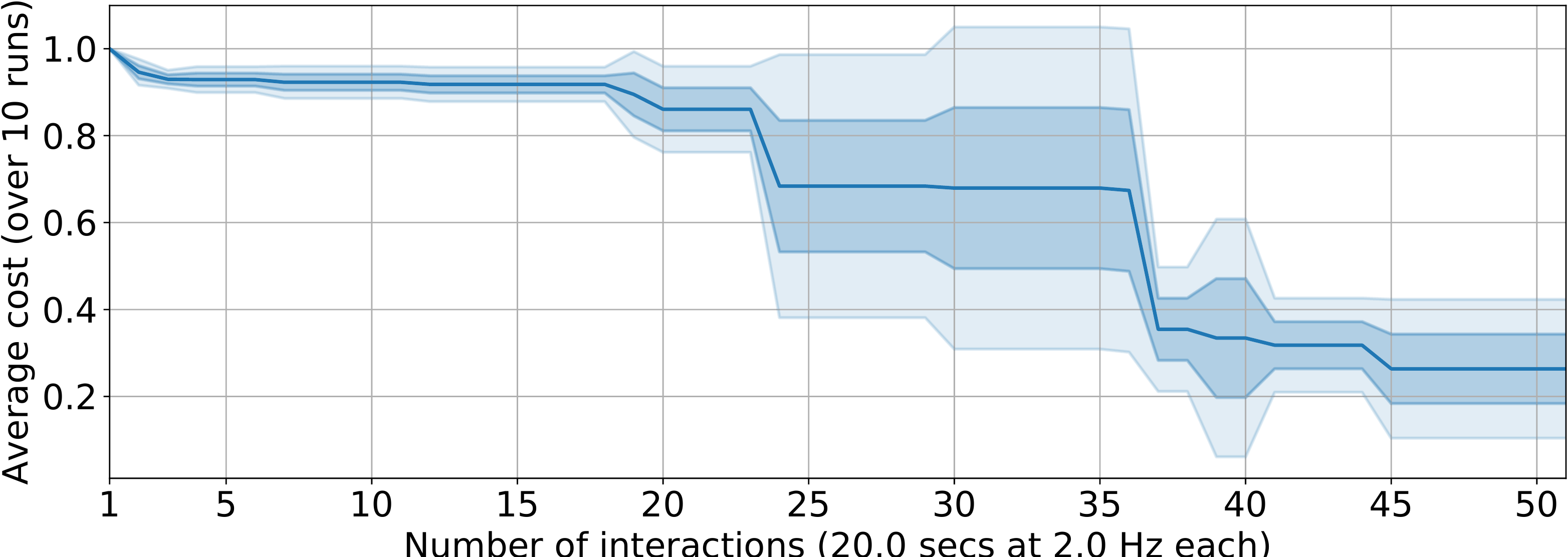}
        \end{subfigure}
        \hfill
        \begin{subfigure}[c]{0.495\textwidth}
            \includegraphics[width=\textwidth]{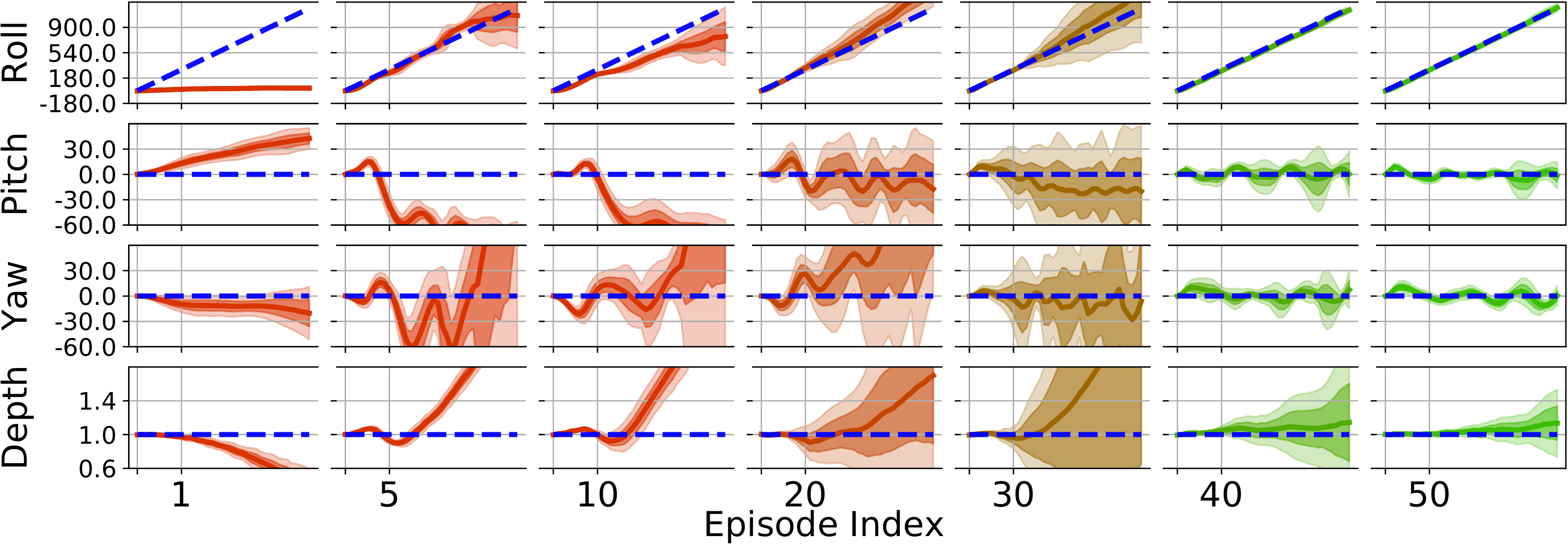}
        \end{subfigure}
        \hfill
        \vspace{-0.25em}
        \caption{\small 6-Leg Corkscrew}
        \vspace{1em}
    \end{subfigure}
    \begin{subfigure}[c]{\textwidth}
        \hfill
        \begin{subfigure}[c]{0.495\textwidth}
            \includegraphics[width=\textwidth]{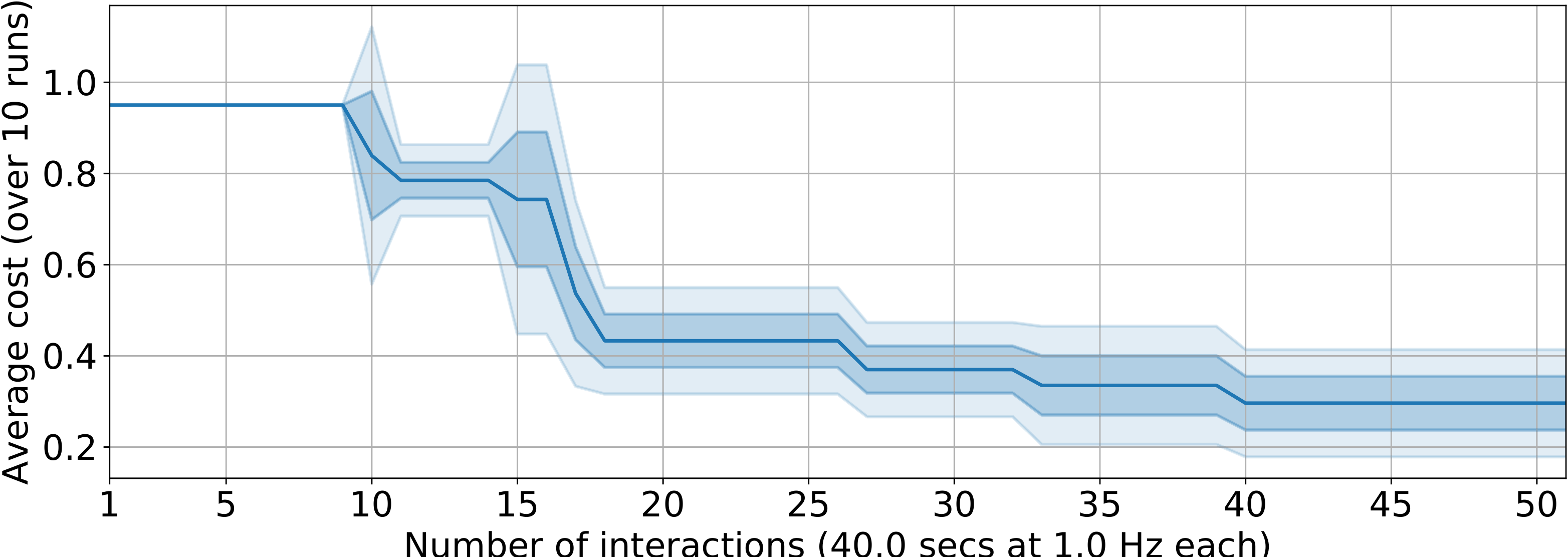}
        \end{subfigure}
        \hfill
        \begin{subfigure}[c]{0.495\textwidth}
            \includegraphics[width=\textwidth]{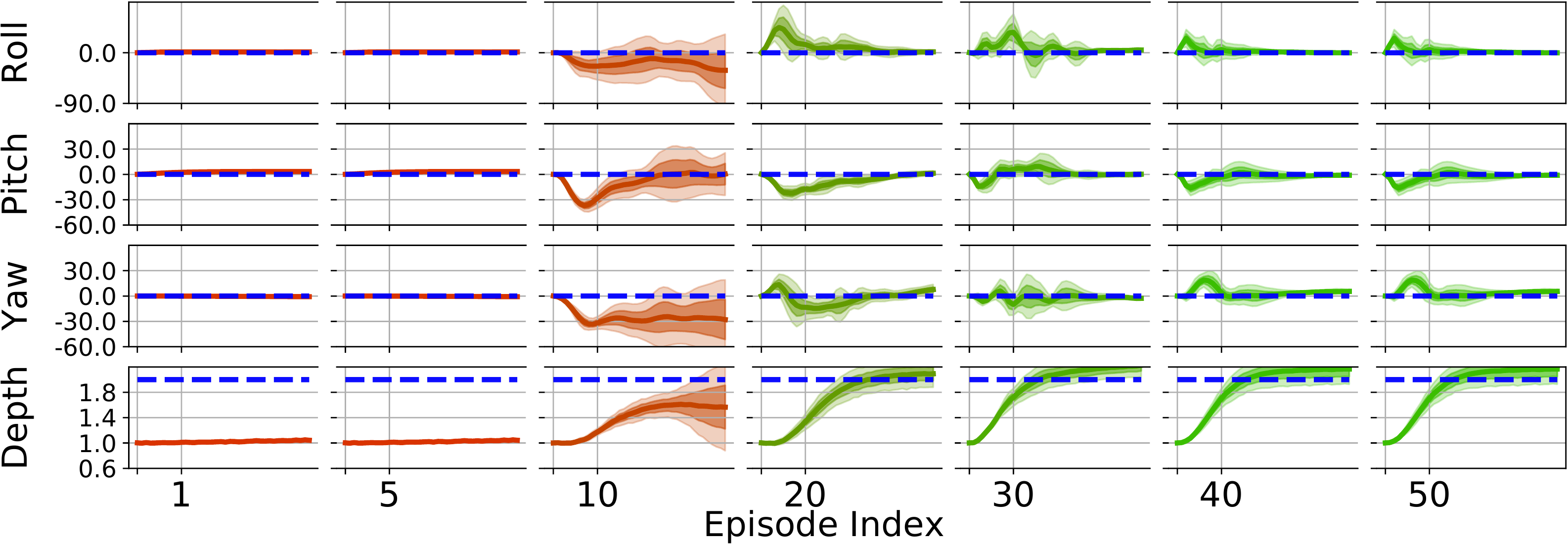}
        \end{subfigure}
        \hfill
        \vspace{-0.25em}
        \caption{\small 6-Leg Depth change}
        \vspace{1em}
    \end{subfigure}
    \vspace{-1em}
     \caption{\small Learning curve and the evolution of the trajectory distribution as learning progresses for 6-leg tasks. In this case, the robot is trying to control the amplitudes, leg angle offsets, and phase offsets for all 6 legs. The algorithm takes longer to converge in this case, when compared to the 2-leg tasks. This is possibly due to the larger state and action spaces (13 state dimensions + 18 action dimensions). Nevertheless, this demonstrates that the algorithm can scale to higher dimensional problems.}
     \label{fig:aqua6leg}
     \vspace{-1.5em}
\end{figure*}
We trained dynamics models and policies with 4 hidden layers of 200 units each. The dynamics models use truncated Log-Normal dropout and we enable dropout for the policy with $p=0.1$. We used a learning rate of $10^{-4}$ and clip gradients to $\nu=1.0$. The experience dataset is initialized with 5 random trials, common to all the tasks with the same state and action spaces.

Fig.~(\ref{fig:aqua2leg}) and~(\ref{fig:aqua6leg}) show the results of gait learning in the simulation environment described in~\cite{meger2015learning}. In addition to learning curves on the left of each task panel, we show detailed state telemetry for selected learning episodes on the right to provide intuition on stability and learning progression. The shaded regions represent the variance of the trajectory distributions on the target system over 10 different runs. In each case attempted, our method was able to learn effective swimming behavior, to coordinate the motions of multiple flippers and overcome simulated hydrodynamic effects without any prior model. For the 2-leg tasks, our method obtains successful policies in 10-20 trials, a number competitive with results reported in~\cite{meger2015learning}. We obtained successful controllers for the depth-change task, which was unsuccessful in prior work. The 6-leg tasks, with their considerably higher-dimensional state and action spaces, take roughly double the number of trials. But all tasks still converged by trial 50, which remains practical for real deployment.

\section{Conclusion}
We have presented improvements to a probabilistic model-based reinforcement learning algorithm, Deep-PILCO, to enable fast synthesis of controllers for robotics applications. Our algorithm is based on treating neural network models trained with dropout as an approximation to the posterior distribution of dynamics models given the experience data. Sampling dynamics models from this distribution helps in avoiding model-bias during policy optimization; policies are optimized for a finite sample of dynamics models, obtained through the application of dropout noise masks. Our changes enable training of neural network controllers, which we demonstrate to outperform RBF controllers on the cart-pole swing-up task. We obtain competitive performance on the task of swing-up and stabilization of a double pendulum on a cart. Finally, we demonstrated the usefulness of the algorithm on the higher dimensional tasks of learning gaits for pose stabilization for a six legged underwater robot. We replicate previous results~\cite{meger2015learning} where we control the robot with 2 flippers, and provide new results on learning to control the robot using all 6 legs, now including phase offsets. 

%While the ability to train deep network dynamics predictors is clearly effective in control, this is by no means the only application possible for this ability within a robot system. We are currently investigating extensions of these basic approaches to adaptive planning and active sensing. The task of active data gathering to optimize self-knowledge has long been a key to effective system identification and model learning. The consistent uncertainty predictions that result from our approach's stable training regime will be an excellent tool to study this active learning problem.
\vspace{-0.75em}
\bibliographystyle{IEEEtran}
\bibliography{IEEEabrv,iros_2018_mbrl_nn_policies}
\end{document}